\pdfoutput=1
\documentclass[11pt]{article}

\usepackage[]{ACL2023}

\usepackage{times}
\usepackage{latexsym}

\usepackage[T1]{fontenc}
\usepackage[utf8]{inputenc}

\usepackage{microtype}
\usepackage{amsmath}
\usepackage{graphicx}
\usepackage{multirow}
\usepackage{soul}

\usepackage{pifont}

\usepackage{times}
\usepackage{latexsym}
\usepackage{graphicx}
\usepackage{booktabs}
\usepackage{subcaption}
\usepackage[ruled,vlined]{algorithm2e}
\usepackage{multirow}
\usepackage{tabularx}
\usepackage{amsmath}
\usepackage{bbold}
\usepackage{mathtools}
\usepackage{xcolor}
\usepackage{booktabs}
\usepackage{tabularx,ragged2e}
\usepackage{enumitem}
\usepackage{scrextend}

\usepackage{inconsolata}

\title{Post-Abstention: Towards Reliably Re-Attempting the Abstained Instances in QA}

\author{Neeraj Varshney and 
  Chitta Baral
  \\
  Arizona State University 
  }

\begin{document}
\maketitle
\begin{abstract}
Despite remarkable progress made in natural language processing, even the state-of-the-art models often make incorrect predictions. 
Such predictions hamper the reliability of systems and limit their widespread adoption in real-world applications.
\textit{Selective prediction} partly addresses the above concern by enabling models to abstain from answering when their predictions are likely to be incorrect.
While selective prediction is advantageous, it leaves us with a pertinent question `\textit{what to do after abstention}'.
To this end, we present an explorative study on `Post-Abstention', a task that allows re-attempting the abstained instances with the aim of increasing \textit{coverage} of the system without significantly sacrificing its \textit{accuracy}.
We first provide mathematical formulation of this task and then explore several methods to solve it.
Comprehensive experiments on $11$ QA datasets show that these methods lead to considerable risk improvements --performance metric of the Post-Abstention task-- both in the in-domain and the out-of-domain settings. 
We also conduct a thorough analysis of these results which further leads to several interesting findings.
Finally, we believe that our work will encourage and facilitate further research in this important area of addressing the reliability of NLP systems.

\end{abstract}

\section{Introduction}

Despite remarkable progress made in Natural Language Processing (NLP), even the state-of-the-art systems often make incorrect predictions. This problem becomes worse when the inputs tend to diverge from the training data distribution \cite{elsahar-galle-2019-annotate,miller2020effect,pmlr-v139-koh21a}. Incorrect predictions hamper the reliability of systems and limit their widespread adoption in real-world applications.

\textbf{Selective prediction} partly addresses the above concern by enabling models to abstain from answering when their predictions are likely to be incorrect.
By avoiding potentially incorrect predictions, it allows maintaining high task accuracy and thus improves the system's reliability.
Selective prediction has recently received considerable attention from the NLP community leading to development of several methods \cite{kamath-etal-2020-selective, garg-moschitti-2021-will, xin-etal-2021-art, varshney-etal-2022-towards}. 
While these contributions are important, selective prediction leaves us with a pertinent question:  \textit{what to do \underline{after abstention}?}

\begin{figure*}
    \centering 
    \small
    \includegraphics[width=0.92\linewidth]{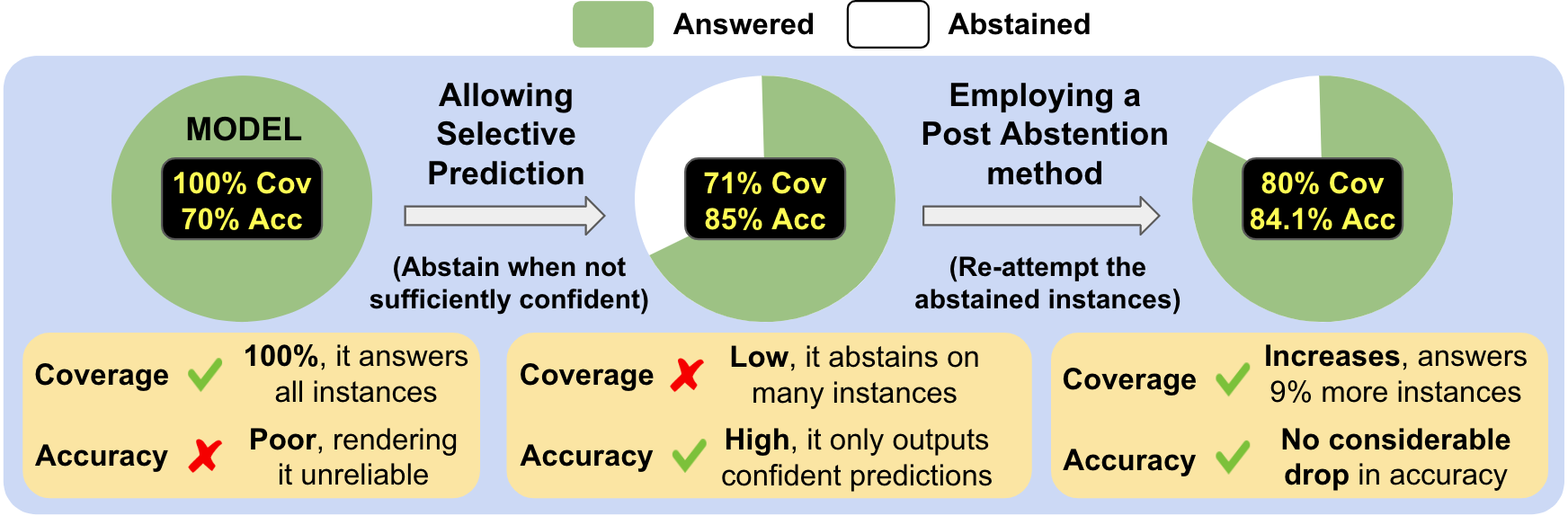}
    \caption{
    Illustrating the \textbf{impact of employing a post-abstention method} on top of selective prediction system. A regular model that has an accuracy of 70\% (at coverage 100\%) is first enabled with selective prediction ability that increases the accuracy to 85\% but drops the coverage to 71\%. Then, on employing a post-abstention method to the abstained instances (remaining 29\%), coverage increases to 80\% without a considerable drop in overall accuracy.
    }
    \label{fig:teaser}
\end{figure*}

In this work, we address the above question and present an explorative study on `\textbf{Post-Abstention}', a task that allows re-attempting the abstained instances with the aim of increasing \textit{coverage} of the given selective prediction system without significantly sacrificing its \textit{accuracy}.
Figure \ref{fig:teaser} illustrates the benefit of employing a post-abstention method; a model that achieves an accuracy of $70\%$ is first enabled with the selective prediction ability that increases the accuracy to $85\%$ but answers only $71\%$ instances. Then, a post-abstention method is employed (for the $29\%$ abstained instances) that assists the system in answering $9\%$ more instances raising the coverage to $80\%$ without considerably dropping the overall accuracy.
We note that this task allows re-attempting all the abstained instances but does not require the system to necessarily output predictions for all of them i.e. the system can abstain even after utilizing a post-abstention method (when it is not sufficiently confident even in its new prediction). 
This facet not only allows the system to maintain its performance but also provides opportunities of sequentially applying stronger post-abstention methods to reliably and optimally increase the coverage in stages.

We provide mathematical formulation of the post-abstention task and explore several baseline methods to solve it (Section \ref{sec_post_abstention}).
To evaluate the efficacy of these methods, we conduct comprehensive experiments with $11$ Question-Answering datasets from MRQA shared task \cite{fisch2019mrqa} in both in-domain and out-of-domain settings (Section \ref{sec_experiments}).
Our post-abstention methods lead to overall risk improvements (performance metric of the proposed task) of up to $21.81$ in the in-domain setting and $24.23$ in the out-of-domain setting.
To further analyze these results, we study several research questions, such as 
`what is the extent of overlap between the instances answered by different post-abstention methods', `what is the distribution of model's original confidence on instances that get answered in the post-abstention stage', and {`how often do the system's predictions change after applying post-abstention methods'}.
In Section \ref{sec_analysis}, we show that these investigations lead to numerous important and interesting findings.

In summary, our contributions are as follows:
\begin{enumerate}[noitemsep,nosep,leftmargin=*]
    \item We present an \textbf{explorative study on `Post-Abstention'}, a task that aims at increasing the \textit{coverage} of a given selective prediction system without significantly sacrificing its \textit{accuracy}.
    
    \item We \textbf{explore several baseline post-abstention methods} and evaluate them in an extensive experimental setup spanning $11$ QA datasets in both in-domain and out-of-domain settings.
    
    \item We show that the proposed post-abstention methods \textbf{result in overall risk value improvements} of up to $21.81$ and $24.23$ in the in-domain and out-of-domain settings respectively.
    
    \item Our \textbf{thorough analysis} leads to several interesting findings, such as 
    (a) instances answered by different post-abstention methods are not mutually exclusive i.e. there exist some overlapping instances,
    (b) instances that get answered in the post-abstention stage are not necessarily the ones on which the given system was initially most confident, etc.
    
\end{enumerate}
We believe our work will encourage further research in Post-Abstention, an important step towards improving the reliability of NLP systems.

\section{Post-Abstention}
\label{sec_post_abstention}
In this section, we first provide background for post-abstention (\ref{subsec_background}) and then describe the task (\ref{subsec_formulation}) and its approaches (\ref{subsec_approaches}).

\subsection{Background}
\label{subsec_background}
Post-abstention, as the name suggests, is applicable for a system that abstains from answering i.e. a selective prediction system.
A system can typically abstain when its prediction is likely to be incorrect.
This improves the reliability of the system.
Such a system typically consists of two functions: a predictor ($f$) that gives the model's prediction on an input ($x$) and a selector ($g$) that determines if the system should output the prediction made by $f$:
\begin{equation*}
        (f,g)(x) =
            \begin{cases}
              f(x), & \text{if g(x) = 1} \\
              Abstain, & \text{if g(x) = 0}
            \end{cases}
\end{equation*}
Typically, $g$ comprises of a prediction confidence estimator $\tilde{g}$ and a threshold $th$ that controls the level of abstention for the system:
\begin{equation*}
    g(x) = \mathbb{1}[\tilde{g}(x)) > th]
\end{equation*}
% where $\mathbb{1}$ is an indicator function.\\
A selective prediction system makes trade-offs between $coverage$ and $risk$.
Coverage at a threshold $th$ is defined as the fraction of total instances answered by the system (where $\tilde{g} > th$) and risk is the error on the answered instances.

With decrease in threshold, coverage will increase, but the risk will usually also increase.
The overall selective prediction performance is measured by the \textit{area under Risk-Coverage curve} \cite{el2010foundations} which plots risk against coverage for all confidence thresholds.
Lower AUC is better as it represents lower average risk across all confidence thresholds.

In NLP, approaches such as Monte-Carlo Dropout \cite{gal2016dropout}, Calibration \cite{kamath-etal-2020-selective,varshney-etal-2022-investigating, varshney-etal-2022-towards, zhang-etal-2021-knowing}, Error Regularization \cite{xin-etal-2021-art} and Label Smoothing \cite{Szegedy2016RethinkingTI} have been studied for selective prediction. In this work, we consider MaxProb \cite{hendrycks17baseline}, a technique that uses the maximum softmax probability across all answer candidates as the confidence estimator. We use this simple technique because the focus of this work is on post-abstention i.e. the next step of selective prediction. However, we note that the task formulation and the proposed methods are general and applicable to all selective prediction approaches. 

\subsection{Task Formulation}
\label{subsec_formulation}

We define the post-abstention task as follows:

\begin{addmargin}[1em]{1em}
\textit{Given a selective prediction system with an abstention threshold, the post-abstention task allows re-attempting the abstained instances with the aim of improving the coverage without considerably degrading the accuracy (or increasing the risk)} of the given system.
\end{addmargin}
Next, we mathematically describe the task and its performance evaluation methodology.

\begin{figure}
    \centering 
    \small
    \includegraphics[width=0.85\linewidth]{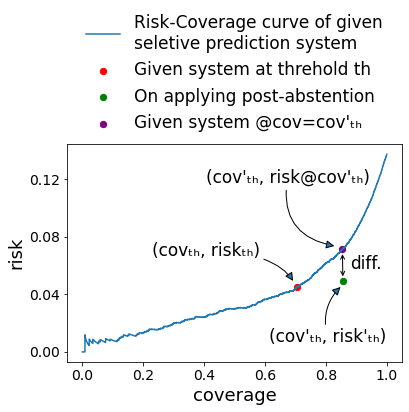}
    \caption{Summarizing \textbf{performance evaluation methodology} of post-abstention. Given a selective prediction system with coverage $cov_{th}$ and risk $risk_{th}$ at abstention threshold $th$, let the new coverage and risk after applying a post-abstention method be $cov'_{th}$ and $risk'_{th}$ respectively. From the risk-coverage curve of the given system, we calculate its risk at coverage $cov'_{th}$ and compare it with $risk'_{th}$ (diff). \textit{For the method to have a positive impact, $risk'_{th}$ should be lower than the risk of the given system at coverage $cov'_{th}$}.
    }
    \label{fig:risk_coverage_curve_performance_metric}
\end{figure}

Let the coverage and risk of the given selective prediction system at abstention threshold $th$ be $cov_{th}$ and $risk_{th}$ respectively. 
A post-abstention method re-attempts the originally abstained instances (where $\tilde{g} < th$) and outputs the new prediction for the ones where it is now sufficiently confident.
This typically leads to an increase in the coverage of the system with some change in the risk value;
let the new coverage and risk be $cov'_{th}$ and $risk'_{th}$ respectively.
From the risk-coverage curve of the given system, we calculate its risk at coverage $cov'_{th}$ and compare it with $risk'_{th}$ to measure the efficacy of the post-abstention method (refer to Figure \ref{fig:risk_coverage_curve_performance_metric}).

\textbf{For a method to have a positive impact}, its risk ($risk'_{th}$) should be lower than the risk of the given system at coverage $cov'_{th}$.
We summarize this performance evaluation methodology in Figure \ref{fig:risk_coverage_curve_performance_metric}.
To get an overall performance estimate of a post-abstention method, we compile these differences in risk values for all confidence thresholds and calculate an aggregated value. The higher the overall improvement value, the more effective the method is.
We note that this evaluation methodology is fair and accurate as it conducts pair-wise comparisons at \textbf{equal coverage} points.
An alternative performance metric could be AUC but it computes the overall area ignoring the pair-wise comparisons which are crucial for our task because the coverage points of the original system would be different from those achieved by the post-abstention method.

\subsection{Approaches}
\label{subsec_approaches}

\subsubsection{Ensembling using Question Paraphrases}
It is well known that even state-of-the-art NLP models are often brittle i.e. when small semantic-preserving changes are made to the input, their predictions tend to fluctuate greatly \cite{jia-liang-2017-adversarial,belinkov2018synthetic,iyyer-etal-2018-adversarial,ribeiro-etal-2018-semantically,wallace-etal-2019-universal}. 
Ensembling the predictions of the model on multiple semantically equivalent variants of the input is a promising approach to address this issue \cite{anantha-etal-2021-open,vakulenko2021comparison} as it can reduce the spread or dispersion of the predictions.

\begin{figure*}
    \centering 
    \small
    \includegraphics[width=0.89\linewidth]{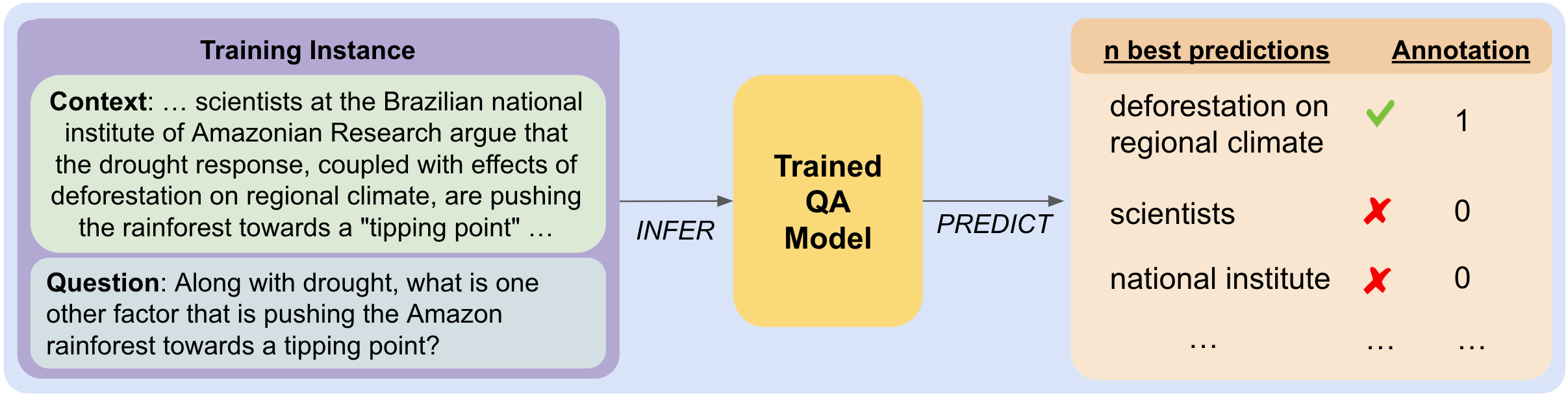}
    \caption{
    Illustrating \textbf{annotation procedure of REToP}. For each training instance, top N predictions given by the QA model are annotated conditioned on their correctness i.e. correct predictions are annotated as `1' and incorrect predictions are annotated as `0'. This annotated binary classification dataset is used to train the auxiliary model.
    }
    \label{fig:approach}
\end{figure*}

We leverage the above technique in re-attempting the abstained questions i.e. we first generate multiple paraphrases of the input instance and then aggregate the model's predictions on them.
We use BART-large \cite{lewis2019bart} model fine-tuned on Quora Question Corpus \cite{iyer2017first}, PAWS \cite{zhang-etal-2019-paws}, and Microsoft Research Paraphrase Corpus \cite{dolan2005automatically} for paraphrasing and explore the following strategies for aggregating the model predictions:
\begin{itemize}[leftmargin=*]
    \item \textbf{Mean}: In this strategy, we calculate the average confidence assigned to each answer candidate across all predictions.
    Then, we select the candidate with the highest average confidence as the system's prediction. Note that the system will output this prediction only if its confidence surpasses the abstention threshold.

    \item \textbf{Max}: Here, like the \textit{mean} strategy, we select the answer candidate with the highest average confidence but we use the maximum confidence assigned to that candidate as its prediction confidence. This is done to push the most confident prediction above the abstention threshold.
    
\end{itemize}

\subsubsection{Re-Examining Top N Predictions (REToP)}

State-of-the-art models have achieved impressive performance on numerous NLP tasks. Even in cases where they fail to make a correct prediction, they are often able to rank the correct answer as one of their top N predictions.
This provides opportunities for re-examining the top N predictions to identify the correct answer in case of abstention. To this end, a model that can estimate the correctness of a prediction can be leveraged.
Following this intuition, we develop an \textbf{auxiliary model} that takes the context, question, and a prediction as input and assigns a score indicating the likelihood of that prediction to be correct.
This model can be used for each of the top N predictions given by the QA model to select the one that is most likely to be the correct answer.

\paragraph{Training Auxiliary Model:}
We first create data instances by annotating (context, question, prediction) triplets conditioned on the correctness of the QA system's predictions and then train a classification model using this data. This model is specific to the given QA system and essentially learns to distinguish its correct and incorrect predictions.

\begin{itemize}[leftmargin=*]
    \item \textbf{Annotate (context, question, prediction) triplets}: We utilize the trained QA model to get its top N predictions for each training instance. Then, we annotate each (context, question, prediction) triplet based on the prediction's correctness i.e. a correct prediction is annotated as `1' and an incorrect prediction is annotated as `0'.
    Figure \ref{fig:approach} illustrates this annotation step.

    \item \textbf{Train a classification model}: Then, a binary classification model is trained using the annotated dataset collected in the previous step. This model specifically learns to distinguish the correct predictions of the QA model from the incorrect ones. 
    Softmax probability assigned to the label `1' corresponds to the likelihood of correctness for each prediction.
    
\end{itemize}
Note that we use the QA model's top N predictions to collect the `0' annotations instead of randomly selecting candidates because this procedure results in highly informative negative instances (that are probable predictions and yet incorrect) and not easy/obvious negatives.
This can help the auxiliary model in learning fine-grained representations distinguishing correct and incorrect predictions.

\paragraph{Leveraging Auxiliary Model:}
For an abstained instance, we compute the likelihood value for each of the top $N$ predictions given by the QA model using our trained auxiliary model.
Then, we calculate the overall confidence ($c$) of each prediction ($p$) as a weighted average of the QA model's probability ($s_q$) and the auxiliary model's likelihood score ($s_a$) i.e. $c_p$ is calculated as:
\begin{equation*}
    c_p = \alpha*s^p_q + (1 - \alpha)*s^p_a
\end{equation*}

where $\alpha$ is a weight parameter.\\
We incorporate QA model's probability as it provides more flexibility to compute the overall confidence. Finally, prediction with the highest overall confidence is selected as the new prediction.
We differentiate this method from existing methods such as calibration in Appendix \ref{sec_differentiating}.

\subsubsection{Human Intervention (HI)}
In intolerant application domains such as biomedicals where incorrect predictions can have serious consequences, human intervention is the most reliable technique to answer the abstained instances.
Human intervention can be in various forms such as providing relevant knowledge to the model, asking clarifying questions \cite{rao-daume-iii-2018-learning} or simplifying the input question.
In this work, we explore a simple human intervention approach in which the system provides multiple predictions instead of only one prediction for the abstained instances. The human can then select the most suitable prediction from the provided predictions.
Performance of this method can be approximated based on the presence of the correct answer in the predictions provided to the human.
Note that the above approach would answer all the abstained instances and hence the coverage would always be 100\%. 
This implies that with the increase in abstention threshold, the risk would monotonically decrease as multiple predictions would be returned for a larger number of instances.

In addition to the above approach, we also explore a \textbf{REToP-centric} HI approach in which the system returns multiple predictions only when REToP surpasses the confidence threshold in the post-abstention stage.
Similar to REToP, it abstains on the remaining instances.
Finally, we note that comparing the performance of HI approaches with other post-abstention approaches would be unfair as other approaches return only a single prediction. Therefore, we present HI results separately.

\begin{table*}[h]
\centering
\resizebox{\textwidth}{!}{%
\begin{tabular}{ll|llllllllllllll|l}
\toprule
\multicolumn{1}{c}{\multirow{2}{*}{\textbf{Dataset}}} &
\multicolumn{1}{c}{\multirow{2}{*}{\textbf{Model}}} &
\multicolumn{2}{c}{\textbf{0.2}} &
  \multicolumn{2}{c}{\textbf{0.32}} &
  \multicolumn{2}{c}{\textbf{0.36}} &
  \multicolumn{2}{c}{\textbf{0.48}} &
  \multicolumn{2}{c}{\textbf{0.54}} &
  \multicolumn{2}{c}{\textbf{0.60}} &
  \multicolumn{2}{c}{\textbf{0.68}} &
  \multicolumn{1}{c}{\textbf{Total Risk}}
  \\
 
 & & Cov$\uparrow$ & Risk$\downarrow$ & Cov$\uparrow$ & Risk$\downarrow$ & Cov$\uparrow$ & Risk$\downarrow$ & Cov$\uparrow$ & Risk$\downarrow$ & Cov$\uparrow$ & Risk$\downarrow$ & Cov$\uparrow$ & Risk$\downarrow$ & Cov$\uparrow$ & Risk$\downarrow$ & \textbf{Improvement$\uparrow$}\\

\midrule
\multirow{1}{*}{\textbf{}} & Given (G) & 

96.65 & 32.45 & 87.24 & 28.10 & 83.34 & 26.69 & 69.94 & 21.91 & 62.57 & 19.91 & 56.23 & 17.98 & 47.92 & 15.43 & 
   \\

\multirow{1}{*}{\textbf{SQuAD}} & REToP & 

99.73 & \textbf{33.75} & 97.27 & \textbf{31.93} & 95.08 & \textbf{30.85} & 80.88 & \textbf{24.84} & 72.44 & \textbf{21.82} & 63.73 & \textbf{19.19} & 52.65 & \textbf{16.43} & 
  \\

(in-domain) & G@REToP$_{cov}$ &     - & 34.00 & - & 32.77 & - & 31.67 & - & 25.82 & - & 22.59 & - & 20.24 & - & 16.83 & \textbf{21.81}

  \\

\midrule
\multirow{3}{*}{\textbf{HotpotQA}} & Given (G) & 97.54 & 67.65 & 89.56 & 65.88 & 85.39 & 65.13 & 71.75 & 62.71 & 64.77 & 61.56 & 58.19 & 60.34 & 49.25 & 58.29 &  \\

& REToP & 99.93 & \textbf{68.17} & 98.63 & \textbf{67.39} & 96.9 & \textbf{66.61} & 82.88 & \textbf{63.61} & 73.55 & \textbf{61.89} & 64.36 & \textbf{60.53} & 52.96 & \textbf{58.34} &    \\

& G@REToP$_{cov}$  &     - & 68.30 & - & 67.92 & - & 67.47 & - & 64.52 & - & 63.04 & - & 61.55 & - & 59.01 & \textbf{21.54}  \\

\midrule
\multirow{3}{*}{\textbf{RE}} & Given (G) &  97.59 & 44.49 & 89.01 & 40.51 & 85.41 & 39.04 & 74.08 & 34.16 & 66.86 & 30.54 & 60.58 & 27.94 & 54.10 & 24.20 &     \\

& REToP & 99.93 & \textbf{45.38} & 98.95 & \textbf{44.39} & 97.52 & \textbf{43.79} & 85.89 & \textbf{38.67} & 77.61 & \textbf{34.57} & 69.54 & \textbf{31.12} & 59.33 & \textbf{25.39} &   \\

& G@REToP$_{cov}$  &     - & 45.47 & - & 45.01 & - & 44.43 & - & 39.22 & - & 35.51 & - & 32.10 & - & 27.33 & \textbf{20.42}\\

\midrule
\multirow{3}{*}{\textbf{RACE}} & Given (G) & 89.02 & 80.5 & 71.07 & 77.04 & 66.17 & 75.56 & 51.34 & 72.54 & 43.47 & 69.62 & 36.2 & 68.85 & 29.97 & 63.86 & 
    \\

& REToP & 99.41 & 82.24 & 92.28 & \textbf{80.71} & 86.94 & \textbf{79.35} & 62.91 & \textbf{73.82} & 51.48 & \textbf{71.76} & 42.28 & \textbf{69.47} & 33.09 & \textbf{65.92} & 
   \\

& G@REToP$_{cov}$  &     - & 81.94 & - & 81.00 & - & 80.00 & - & 75.00 & - & 72.54 & - & 69.72 & - & 66.37 & \textbf{15.10} \\

\midrule
\multirow{3}{*}{\textbf{NewsQA}} & Given (G) &  93.90 & 69.76 & 80.91 & 66.40 & 75.5 & 64.91 & 60.30 & 60.79 & 53.30 & 58.8 & 47.17 & 56.62 & 39.32 & 54.11 &     \\

& REToP & 
99.48 & \textbf{71.03} & 96.13 & \textbf{70.24} & 93.21 & 69.64 & 70.85 & \textbf{63.71} & 60.73 & \textbf{60.67} & 52.04 & \textbf{58.07} & 42.09 & \textbf{54.94} & 
  \\

& G@REToP$_{cov}$  &     - & 71.31 & - & 70.36 & - & 69.61 & - & 63.81 & - & 61.01 & - & 58.33 & - & 55.02 & \textbf{5.10}
 \\

\midrule
\multirow{3}{*}{\textbf{SearchQA}} & Given (G) &  96.15 & 86.68 & 81.77 & 85.67 & 75.77 & 85.34 & 58.64 & 84.08 & 50.22 & 83.58 & 42.67 & 83.33 & 34.46 & 82.55 & \\

& REToP & 99.92 & 87.06 & 97.58 & 86.81 & 93.92 & \textbf{86.48} & 71.49 & \textbf{84.76} & 59.46 & \textbf{84.04} & 48.6 & \textbf{83.48} & 37.08 & \textbf{82.75} &  \\

& G@REToP$_{cov}$  &     - & 87.04 & - & 86.79 & - & 86.52 & - & 85.07 & - & 84.15 & - & 83.56 & - & 82.77 & \textbf{1.78}  \\

\midrule
\multirow{3}{*}{\textbf{TriviaQA}} & Given (G) & 96.67 & 67.31 & 86.89 & 65.05 & 82.54 & 63.82 & 68.81 & 60.39 & 61.44 & 58.39 & 55.11 & 56.48 & 47.12 & 54.03 & 
   \\

& REToP & 99.86 & \textbf{68.07} & 97.07 & \textbf{67.33} & 93.72 & \textbf{66.23} & 76.72 & 62.40 & 67.93 & 60.25 & 59.55 & \textbf{57.77} & 49.29 & 54.89 & 
  \\

& G@REToP$_{cov}$  &     - & 68.09 & - & 67.42 & - & 66.60 & - & 62.32 & - & 60.12 & - & 57.95 & - & 54.83 & \textbf{0.70}
  \\

\midrule
\multirow{3}{*}{\textbf{NQ}} & Given (G) & 92.37 & 63.78 & 79.04 & 59.99 & 74.87 & 58.77 & 60.60 & 53.51 & 54.03 & 51.00 & 47.94 & 48.31 & 41.70 & 45.27 & 
  \\

& REToP & 98.71 & \textbf{65.34} & 93.04 & \textbf{63.39} & 89.30 & \textbf{62.62} & 70.65 & \textbf{56.90} & 61.68 & \textbf{53.54} & 53.24 & \textbf{50.10} & 43.75 & \textbf{46.44} & 
  \\

& G@REToP$_{cov}$  &     - & 65.67 & - & 63.93 & - & 63.02 & - & 57.43 & - & 53.80 & - & 50.68 & - & 46.45 & \textbf{10.70}  \\

\midrule
\multirow{3}{*}{\textbf{DROP}} & Given (G) & 95.74 & 88.46 & 81.17 & 87.38 & 76.11 & 87.33 & 62.34 & 86.23 & 53.69 & 85.38 & 48.77 & 84.45 & 43.05 & 85.01 &   \\

& REToP & 99.53 & 88.64 & 92.95 & \textbf{87.83} & 88.42 & 88.04 & 69.00 & \textbf{86.31} & 58.55 & \textbf{85.57} & 51.90 & \textbf{84.49} & 44.18 & 85.09 &   \\

& G@REToP$_{cov}$  &     - & 88.63 & - & 88.19 & - & 87.88 & - & 86.69 & - & 85.91 & - & 84.87 & - & 84.94 & \textbf{3.63}
  \\

\midrule
\multirow{3}{*}{\textbf{DuoRC}} & Given (G) &  97.20 & 68.68 & 87.87 & 66.41 & 84.21 & 65.82 & 71.09 & 62.42 & 64.16 & 61.47 & 57.16 & 59.91 & 50.03 & 58.46 &   \\

& REToP & 99.87 & \textbf{69.45} & 98.33 & \textbf{69.17} & 96.14 & 68.68 & 80.75 & \textbf{64.69} & 71.95 & \textbf{62.59} & 62.56 & \textbf{60.70} & 52.90 & \textbf{58.69} &  \\

& Original@cov &     - & 69.51 & - & 69.02 & - & 68.4 & - & 64.77 & - & 62.74 & - & 60.92 & - & 59.32 & \textbf{4.32} \\

\midrule
\multirow{3}{*}{\textbf{TBQA}} & Given (G) &  94.34 & 67.14 & 80.9 & 63.32 & 75.65 & 61.92 & 57.49 & 56.02 & 49.63 & 52.14 & 41.45 & 51.04 & 34.07 & 50.00 & \\

& REToP & 99.53 & \textbf{68.38} & 95.01 & \textbf{67.23} & 91.68 & \textbf{66.18} & 68.20 & \textbf{58.34} & 58.55 & \textbf{54.77} & 47.37 & \textbf{51.26} & 37.26 & \textbf{49.64} &  \\

& G@REToP$_{cov}$  &     - & 68.56 & - & 67.30 & - & 66.23 & - & 59.41 & - & 56.02 & - & 52.60 & - & 50.71 & \textbf{24.23} \\

\bottomrule
\end{tabular}
}
\caption{\textbf{Performance of REToP as a post-abstention method} for selected abstention thresholds. 
The QA model is trained using SQuAD training data and evaluated on SQuAD (in-domain) and 10 out-of-domain datasets.
For each dataset, we provide three rows:
first row (`\textit{Given}') shows the coverage and risk values of the given selective prediction system at different abstention thresholds,
second row (`\textit{REToP}') shows the coverage and risk after applying REToP on abstained instances of the given system, and
third row (`\textit{G@REToP}$_{cov}$') shows risk of the given system at the coverage achieved by \textit{REToP}.
For the post abstention method to be effective, risk in the second row should be less than that in the third row and the magnitude of difference corresponds to the improvement.
The last column corresponds to the overall improvement aggregated over all confidences ranging from 0 to 1 at an interval of 0.02.
$\downarrow$ and $\uparrow$ indicate that lower (risk) and higher (coverage, risk improvement) values are better respectively.
}
\label{tab:REToP_perf}
\end{table*}
\section{Experiments and Results}
\label{sec_experiments}
\subsection{Experimental Setup}
\paragraph{Datasets:}
We experiment with SQuAD 1.1 \cite{rajpurkar-etal-2016-squad} as the source dataset and the following $10$ datasets as out-of-domain datasets: NewsQA \cite{trischler-etal-2017-newsqa}, TriviaQA \cite{joshi-etal-2017-triviaqa}, SearchQA \cite{dunn2017searchqa}, HotpotQA \cite{yang-etal-2018-hotpotqa}, and Natural
Questions \cite{kwiatkowski-etal-2019-natural}, DROP \cite{dua-etal-2019-drop}, DuoRC \cite{saha-etal-2018-duorc}, RACE \cite{lai-etal-2017-race}, RelationExtraction \cite{levy-etal-2017-zero}, and TextbookQA \cite{kim-etal-2019-textbook}.
We use the preprocessed data from the MRQA shared task \cite{fisch2019mrqa} for our experiments.

\paragraph{Implementation Details: }
We run all our experiments using the huggingface \cite{wolf-etal-2020-transformers} implementation of transformers on Nvidia V100 16GB GPUs with a batch size of $32$ and learning rate ranging in $\{1{-}5\}e{-}5$. 
We generate $10$ paraphrases of the question in Ensembling method, re-examine top $10$ predictions, vary $\alpha$ in the range $0.3-0.7$ for REToP method, and vary the number of predictions in the range $2$ to $5$ for HI methods.
Since the focus of this work is on post-abstention, it's crucial to experiment with models that leave sufficient room for effectively evaluating the ability of post-abstention methods. 
For that reason, we experiment with a small size model (BERT-mini having just 11.3M parameters) from \citet{turc2019well} for our experiments.
However, we note that our methods are general and applicable for all models.

\subsection{Results}

\subsubsection{REToP}
Table \ref{tab:REToP_perf} shows the post-abstention performance of REToP for selected abstention thresholds. 
The last column (`\textit{Total Risk Improvement}') in this table corresponds to the overall improvement aggregated over all confidence thresholds.
It can be observed that REToP achieves considerable risk improvements both in the in-domain setting ($21.81$ on SQuAD) and the out-of-domain settings ($24.23$ on TextbookQA, $21.54$ on HotpotQA, $20.42$ on RE, etc). 
Next, we analyze these results in detail.

\paragraph{Higher improvement on moderate confidences:}

\begin{figure}[t!]
\centering

    \begin{subfigure}{.5\linewidth}
        \includegraphics[width=\linewidth]{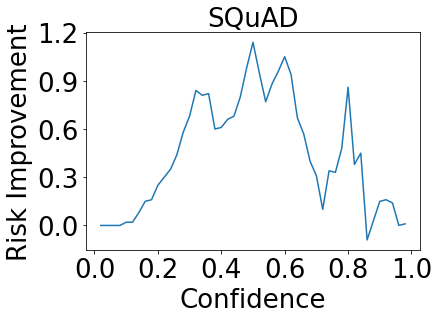}
    \end{subfigure}
    \begin{subfigure}{.43\linewidth}
         \includegraphics[width=\linewidth]{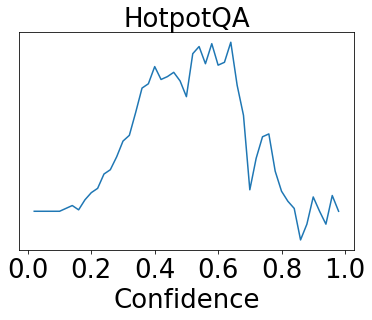}
    \end{subfigure}
    \caption{\textbf{Improvement in risk} achieved by using REToP in post-abstention on SQuAD (in-domain) and HotpotQA (out-of-domain) datasets for all confidences.}
    \label{fig:confidence_vs_improvement}
\end{figure}

In Figure \ref{fig:confidence_vs_improvement}, we plot risk improvements achieved by REToP on SQuAD (in-domain) and HotpotQA (out-of-domain) datasets for all confidence thresholds. 
These plots reveal that the improvement is more on moderate thresholds as compared to low thresholds. 
We attribute this to the high difficulty of instances that remain to be re-attempted at low thresholds i.e. only the instances on which the given system was highly underconfident are left for the post-abstention method. 
It has been shown that model's confidence is negatively correlated with difficulty \cite{swayamdipta-etal-2020-dataset, rodriguez-etal-2021-evaluation, varshney-etal-2022-ildae} implying that the remaining instances are tough to be answered correctly.
This justifies the lesser improvement in performance observed at low thresholds.

\paragraph{In-Domain vs Out-of-Domain Improvement:}
REToP achieves higher performance improvement on the in-domain dataset than the out-of-domain datasets (on average). 
This is expected as the auxiliary model in REToP is trained using the in-domain training data.
However, it still has good performance on out-of-domain datasets as the auxiliary model learns fine-grained representations to distinguish between correct and incorrect predictions.
Furthermore, the improvement on out-of-domain data varies greatly across datasets (from 0.7 on TriviaQA to 24.23 on TextbookQA).

\subsubsection{Comparing Post-Abstention Approaches}
We provide the performance tables for other post-abstention approaches in Appendix.
However, we compare their total risk improvement values in Table \ref{tab:perf_improvement}.
In the in-domain setting, REToP achieves higher improvement than Ensembling method. 
This is because the auxiliary model in REToP has specifically learned to distinguish the correct and incorrect predictions from the training data of this domain.
However, in some out-of-domain cases, Ensembling outperforms REToP (SearchQA, TriviaQA, NewsQA).
Overall, REToP leads to a consistent and higher risk improvement on average.
Ensembling also leads to a minor degradation in a few out-of-domain datasets (DuoRC and TextbookQA).
Next, we analyze the performance of human intervention (HI) methods.

\begin{table}[t]
    \centering
    \small 
    \resizebox{\linewidth}{!}{
    \begin{tabular}{l|lll|l}
    \toprule
        \textbf{Dataset} 
        & \textbf{Ens.} & \textbf{REToP} & \textbf{REToP} & \textbf{*HI on}\\
          & & ($\alpha=0.6$) & ($\alpha=0.65$) & (REToP)\\
        \midrule
        \textbf{SQuAD} & 0.29  & 21.81 & 20.02 & 47.85 \\ 
        \midrule
        \textbf{HotpotQA} & 0.93  & 21.54 & 19.00  & 37.88\\ 
        \textbf{RE} & 21.72  & 20.42 & 17.61  & 46.65 \\ 
        \textbf{RACE} & 16.72  & 15.10 & 14.17  & 36.26 \\ 
        \textbf{NewsQA} & 11.92  & 5.10 & 5.10  &  26.41 \\ 
        \textbf{SearchQA} & 17.05  & 1.78 & 2.23  & 20.08 \\ 
        \textbf{TriviaQA} & 9.50  & 0.70 & 1.47  & 17.21 \\ 
        \textbf{NQ} & 13.40 &   10.70 & 10.89  & 31.95 \\ 
        \textbf{DROP} & 1.57 &   3.63 & 2.99  & 8.08 \\ 
        \textbf{DuoRC} & -1.69   & 4.32 & 5.90  & 20.26 \\ 
        \textbf{TBQA} & -6.93   & 24.23 & 23.73  & 45.18 \\ 
        \midrule
        \textbf{Total} & 84.48   & \textbf{129.33} & 123.11  & 337.81 \\

    \bottomrule
    \end{tabular}
    }
    \caption{\textbf{Comparing total risk improvement} achieved by different post-abstention methods.
    * for HI indicates that it's results are not directly comparable as it outputs multiple predictions while others output only one.}
    \label{tab:perf_improvement}
\end{table}

\subsubsection{Human Intervention (HI)}
We study two variants of HI method. 
In the first variant, multiple predictions (n=2) are returned for all the abstained instances. This makes the coverage to be 100\% for all the confidences; therefore, we present only the risk values in Table \ref{tab:human_intervention_topn}. 
As expected, with increase in abstention threshold, the risk decreases because multiple predictions get outputted for a larger number of instances.
Selection of operating threshold for an application depends on the trade-off between risk that can be tolerated and human effort required to select the most suitable prediction from a set of predictions returned by the system.
For example, a low threshold can be selected for tolerant applications like movie recommendations and a high threshold for tolerant applications like house robots. 

In the second variant of HI method, we study a \textbf{REToP-centric} approach in which the system returns multiple predictions only when REToP surpasses the confidence threshold in the post-abstention stage. 
The last column in Table \ref{tab:perf_improvement} shows the risk improvements achieved by this approach (n=2).
Note that REToP re-examines the top N predictions and selects one while this method outputs multiple predictions and requires a human to select the most suitable one. 
These results indicate that though REToP achieves good performance, there is still some room for improvement.

\begin{table}[t]
    \centering
    \small 
    \resizebox{\linewidth}{!}{
    \begin{tabular}{l|lllll}
    \toprule
        \textbf{Dataset} & \textbf{0.0}  & \textbf{0.2} 
        & \textbf{0.4} & \textbf{0.6} & \textbf{0.8}\\
        \midrule

        \textbf{SQuAD} & 34.15 & 33.72 & 30.9 & 28.05  & 26.3 \\
        \midrule
        \textbf{HotpotQA} & 68.33 & 68.19 & 66.56 & 63.65  & 61.57 \\
        
        \textbf{RE} & 45.52 & 45.35 & 43.39 & 41.28  & 39.31 \\
        
        \textbf{RACE} & 82.05 & 81.6 & 80.12 & 78.19  & 77.15 \\

        \textbf{NewsQA} & 71.46 & 71.2 & 69.42 & 67.21  & 65.29 \\
        
        \textbf{SearchQA} & 87.06 & 86.92 & 85.64 & 83.98  & 82.94 \\
        
        \textbf{TriviaQA} & 68.13 & 67.9 & 66.62 & 64.21  & 62.47 \\

        \textbf{NQ} & 66.09 & 65.67 & 63.63 & 61.06  & 59.31 \\
        
        \textbf{DROP} & 88.69 & 88.69 & 87.56 & 86.36  & 85.7 \\
        
        \textbf{DuoRC} & 69.55 & 69.42 & 68.15 & 66.42  & 65.22 \\

        \textbf{TBQA} & 68.73 & 68.46 & 67.07 & 64.74  & 64.01 \\

    \bottomrule
    \end{tabular}
    }
    \caption{Comparing \textbf{risk values achieved by the HI method} (returns two predictions for all abstained instances) across different abstention thresholds. 
    }
    
    \label{tab:human_intervention_topn}
\end{table}

\subsubsection{Ensembling Using Paraphrases}
Comparing the performance of Mean and Max Ensembling strategies reveals that Max increases the coverage more than the Mean strategy but it also increases the risk considerably.
Thus, pushing the instance's confidence to surpass the abstention threshold fails to provide risk improvements.
However, such a technique could be employed in scenarios where risk degradation can be tolerated.

\section{Analysis}
\label{sec_analysis}
\paragraph{What is the distribution of model's original confidence on the instances that get answered after applying post-abstention method?}

\begin{figure}
    \centering 
    \small
    \includegraphics[width=0.8\linewidth]{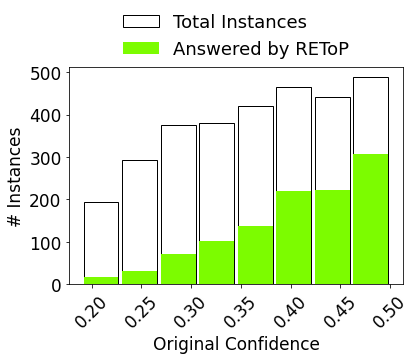}
    \caption{Distribution of QA model's confidence on SQuAD instances that get answered after applying REToP at abstention threshold $0.5$. 
    }
    \label{fig:rq_1}
\end{figure}

In Figure \ref{fig:rq_1}, we show the distribution of model's original confidence on SQuAD instances that get answered by REToP at abstention threshold 0.5. 
Green-colored bars represent the number of instances answered from each confidence bucket.
\textit{We found that REToP answers a large number of instances from the high confidence buckets; however, instances from even low confidence buckets get answered.}
This can further be controlled using the weight parameter ($\alpha$) in the overall confidence computation.

\begin{figure}
    \centering 
    \small
    \includegraphics[width=0.60\linewidth]{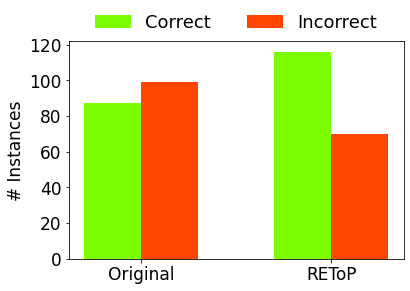}
    \caption{Number of correct (green) and incorrect (red) predictions on those abstained SQuAD instances where REToP surpasses the abstention threshold of 0.5 but its prediction differs from the original system.}
    
    \label{fig:rq_2}
\end{figure}

\paragraph{How often do the system's predictions change after applying REToP and what is its impact?}
REToP can either boost the confidence of the top most prediction of the given model or can select a different answer by re-examining its top N predictions.
In Figure \ref{fig:rq_2}, we specifically analyze the latter scenario i.e. the instances on which REToP's prediction differs from the original model's prediction.
At a threshold of $0.5$, the original system abstains on $3411$ SQuAD instances and after applying REToP, it answers $1110$ of those instances. 
Out of these $1110$ instances, the REToP changes the prediction on $186$ instances. 
The original prediction is incorrect in more cases ($99$ vs $87$) and after applying REToP, the system gives $116$ correct predictions and only $70$ incorrect. 
This implies that by overriding the original system's prediction, REToP improves the system's accuracy. 
However, in some cases, it also changed a correct prediction to incorrect but such cases are lesser than the former.

\begin{figure}
    \centering 
    \small
    \includegraphics[width=0.5\linewidth]{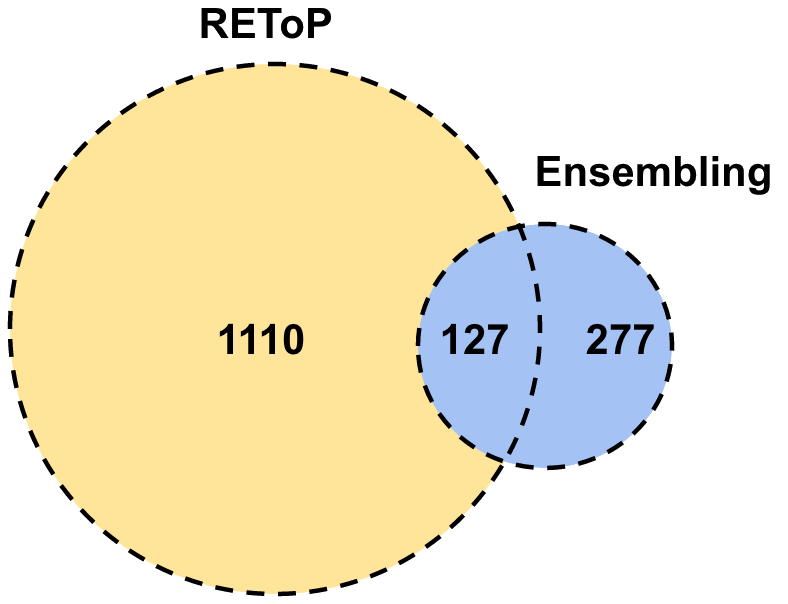}
    \caption{Venn diagram of abstained SQuAD instances answered by REToP and Ensembling (Mean) approaches at abstention threshold $0.5$.
    }
    
    \label{fig:rq_3}
\end{figure}

\paragraph{To what extent do the instances answered by different post-abstention methods overlap?}
In Figure \ref{fig:rq_3}, we demonstrate the Venn diagram of SQuAD instances answered by REToP and Ensembling (Mean) approaches at abstention threshold $0.5$.
REToP answers $1110$ instances while Ensembling answers $277$ and there $127$ common instances between the two approaches.
This indicates that the two sets are not mutually exclusive i.e. there are some instances that get targeted by both the approaches; however, there are a significant number of instances that are not in the intersection.
This result motivates studying composite or sequential application of different post-abstention methods to further improve the post-abstention performance.


\section{Conclusion and Discussion}
In this work, we formulated `Post-Abstention', a task that allows re-attempting the abstained instances of the given selective prediction system with the aim of increasing its \textit{coverage} without significantly sacrificing the \textit{accuracy}.
We also explored several baseline methods for this task.
Through comprehensive experiments on $11$ QA datasets, we showed that these methods lead to considerable performance improvements in both in-domain and out-of-domain settings. We further performed a thorough analysis that resulted in several interesting findings.

Looking forward, we believe that our work opens up several avenues for new research, such as exploring \textit{test-time adaptation}, \textit{knowledge hunting}, and other human intervention techniques like \textit{asking clarification questions} as post-abstention methods (discussed in Appendix \ref{sec_appendix_dicussion}). Studying the impact of composite or sequential application of multiple post-abstention methods in another promising direction. 
Furthermore, prior selective prediction methods can also be repurposed and explored for this task. 
We plan to pursue these crucial research directions in our future work.
Finally, we hope our work will encourage further research in this important area and facilitate the development of more reliable NLP systems.

\section*{Limitations}
The proposed post-abstention methods require additional computation and storage. 
Despite this additional requirement, we note that this is not a serious concern as current devices have high storage capacity and computation hardware. 
Furthermore, additional computation for training auxiliary model in REToP is required only once and just an inference is required at evaluation time which has a much lower computation cost.
Moreover, the risk mitigation that comes with the post-abstention methods weighs much more than the computational or storage overhead in terms of importance.
Secondly, human-intervention techniques require a human to be a participant and contribute in the answering process. 
However, these approaches do not expect the participating human to be an expert in the task.
Like other empirical research, it is difficult to exactly predict the magnitude of improvement a post-abstention method can bring. Our idea of exploring sequential application of multiple post-abstention methods addresses this concern and can be used based on the application requirements.

\section*{Acknowledgement}
We thank the anonymous reviewers for their insightful feedback. This research was supported by DARPA SAIL-ON program.

\bibliography{anthology,custom}
\bibliographystyle{acl_natbib}

\appendix

\section*{Appendix}
\section{Ensembling (Mean) Performance}
\label{sec:ensembling_perf}

Table \ref{tab:Ensembling_perf} shows the performance of using Ensembling (Mean) as a post-abstention method for a few selected abstention threshold values. 
For each dataset, we provide three rows: the first row (`\textit{Given}') shows the coverage and risk values of the given selective prediction system at specified abstention thresholds, 
the second row (`\textit{Ens}') shows the coverage and risk after applying the post-abstention method on the abstained instances of the given selective prediction system, and the final row (`\textit{G@Ens}$_{cov}$') shows the risk of the given selective system at the coverage achieved by \textit{Ens} method.
For the post-abstention method to be effective the risk in the second row should be less than that in the third row and the magnitude of difference corresponds to the improvement.
The last column `\textit{Total Risk Improvement}' shows the overall improvement aggregated over all confidence thresholds ranging between 0 and 1 at an interval of 0.02.

\section{Dataset Statistics}

Table \ref{tab:dataset_stats} shows the statistics of all evaluation datasets used in this work. 
SQuAD corresponds to the in-domain dataset while the remaining 10 datasets are out-of-domain.
We use the pre-processed data from the MRQA shared task \cite{fisch2019mrqa}.

\section{Differentiating REToP from Calibration}
\label{sec_differentiating}
REToP is different from calibration based techniques presented in \cite{kamath-etal-2020-selective, varshney-etal-2022-investigating} in the following aspects: \\
(a) Firstly, REToP does not require a held-out dataset unlike calibration based methods that infer the model on the held-out dataset to gather instances on which the model in incorrect.\\
(b) Secondly, the auxiliary model trained in REToP predicts the likelihood of correctness of (context, question, prediction) triplet i.e. it is used for each of the top N prediction individually. This is in contrast to calibrators that predicts a single score for an instance and ignores the top N predictions.\\
(c) Finally, we use the entire context, question, and the prediction to predict its correctness likelihood score unlike feature-based calibrator models in which a random-forest model is trained using just syntax-level features such as length of question, semantic similarity of prediction with the question, etc.
\begin{table}
\small
    \centering
    \begin{tabular}{p{1.7cm}p{1cm}|p{1.8cm}p{0.9cm}}
     \toprule
        \textbf{Dataset} & \textbf{Size} & \textbf{Dataset} & \textbf{Size}\\
         \midrule
        SQuAD                &    10507 & HotpotQA             &     5901 \\
        %  \midrule
        RE   &     2948  & RACE                 &      674 \\ 
        % \midrule
        NewsQA               &     4212 & SearchQA        &         16980 \\ 
        % \midrule
        TriviaQA         &     7785  & NQ  &  12836  \\ 
        % \midrule
        DROP                 &     1503 & DuoRC    &    1501\\ 
        % \midrule
        TBQA           &     1503  \\ 
        \bottomrule
    \end{tabular}
    \caption{Statistics of evaluation data used in this work.}
    \label{tab:dataset_stats}
\end{table}

\begin{table*}[h]
\centering
\resizebox{\textwidth}{!}{%
\begin{tabular}{ll|llllllllllllll|l}
\toprule
\multicolumn{1}{c}{\multirow{2}{*}{\textbf{Dataset}}} &
\multicolumn{1}{c}{\multirow{2}{*}{\textbf{Model}}} &
\multicolumn{2}{c}{\textbf{0.2}} &
  \multicolumn{2}{c}{\textbf{0.32}} &
  \multicolumn{2}{c}{\textbf{0.36}} &
  \multicolumn{2}{c}{\textbf{0.48}} &
  \multicolumn{2}{c}{\textbf{0.54}} &
  \multicolumn{2}{c}{\textbf{0.60}} &
  \multicolumn{2}{c}{\textbf{0.68}} &
  \multicolumn{1}{c}{\textbf{Total Risk}}
  \\
 
 & & Cov$\uparrow$ & Risk$\downarrow$ & Cov$\uparrow$ & Risk$\downarrow$ & Cov$\uparrow$ & Risk$\downarrow$ & Cov$\uparrow$ & Risk$\downarrow$ & Cov$\uparrow$ & Risk$\downarrow$ & Cov$\uparrow$ & Risk$\downarrow$ & Cov$\uparrow$ & Risk$\downarrow$ & \textbf{Improvement$\uparrow$}\\

\midrule
\multirow{1}{*}{\textbf{}} & Given (G) & 

96.65 & 32.45 & 87.24 & 28.10 & 83.34 & 26.69 & 69.94 & 21.91 & 62.57 & 19.91 & 56.23 & 17.98 & 47.92 & 15.43 & 
   \\

\multirow{1}{*}{\textbf{SQuAD}} & Ens & 

97.64 & 32.88 & 89.51 & 28.93 & 87.64 & 28.24 & 72.46 & 22.71 & 65.12 & 20.58 & 58.37 & 18.7 & 49.59 & 15.89 &
  \\

(in-domain) & G@Ens$_{cov}$ &     - & 32.96 & - & 29.09 & - & 28.26 & - & 22.58 & - & 20.65 & - & 18.66 & - & 15.91 & 0.29

  \\

\midrule
\multirow{3}{*}{\textbf{HotpotQA}} & Given (G) & 97.54 & 67.65 & 89.56 & 65.88 & 85.39 & 65.13 & 71.75 & 62.71 & 64.77 & 61.56 & 58.19 & 60.34 & 49.25 & 58.29 &  \\

& Ens & 98.59 & 67.84 & 91.93 & 66.23 & 90.41 & 65.92 & 75.65 & 63.17 & 68.45 & 62.22 & 61.31 & 60.72 & 52.26 & 58.88 &    \\

& G@Ens$_{cov}$  &     - & 67.9 & - & 66.37 & - & 66.04 & - & 63.4 & - & 62.14 & - & 60.91 & - & 58.94 & 0.93 \\

\midrule
\multirow{3}{*}{\textbf{RE}} & Given (G) &  97.59 & 44.49 & 89.01 & 40.51 & 85.41 & 39.04 & 74.08 & 34.16 & 66.86 & 30.54 & 60.58 & 27.94 & 54.10 & 24.20 &     \\

& Ens & 98.27 & 44.56 & 92.2 & 41.35 & 90.57 & 40.71 & 77.44 & 34.87 & 70.86 & 31.45 & 64.86 & 29.08 & 56.07 & 24.74  &   \\

& G@Ens$_{cov}$  &     - & 44.82 & - & 42.27 & - & 41.42 & - & 35.58 & - & 32.47 & - & 30.02 & - & 25.54 & 21.72 \\

\midrule
\multirow{3}{*}{\textbf{RACE}} & Given (G) & 89.02 & 80.5 & 71.07 & 77.04 & 66.17 & 75.56 & 51.34 & 72.54 & 43.47 & 69.62 & 36.2 & 68.85 & 29.97 & 63.86 & 
    \\

& Ens & 91.69 & 80.42 & 73.89 & 77.71 & 71.51 & 77.18 & 53.71 & 72.65 & 46.88 & 70.25 & 40.21 & 69.0 & 31.6 & 64.79  & 
   \\

& G@Ens$_{cov}$  &     - & 80.88 & - & 77.31 & - & 77.13 & - & 72.93 & - & 71.43 & - & 70.11 & - & 65.09 & 16.72 \\

\midrule
\multirow{3}{*}{\textbf{NewsQA}} & Given (G) &  93.90 & 69.76 & 80.91 & 66.40 & 75.5 & 64.91 & 60.30 & 60.79 & 53.30 & 58.8 & 47.17 & 56.62 & 39.32 & 54.11 &     \\

& Ens & 
95.56 & 70.24 & 83.52 & 67.14 & 81.13 & 66.49 & 63.01 & 61.53 & 55.75 & 59.45 & 49.53 & 57.19 & 41.17 & 54.21 & 
  \\

& G@Ens$_{cov}$  &     - & 70.18 & - & 67.02 & - & 66.46 & - & 61.63 & - & 59.67 & - & 57.33 & - & 54.67 & 11.92
 \\

\midrule
\multirow{3}{*}{\textbf{SearchQA}} & Given (G) &  96.15 & 86.68 & 81.77 & 85.67 & 75.77 & 85.34 & 58.64 & 84.08 & 50.22 & 83.58 & 42.67 & 83.33 & 34.46 & 82.55 & \\

& Ens & 98.0 & 86.82 & 87.31 & 85.79 & 84.7 & 85.61 & 65.65 & 84.1 & 56.86 & 83.65 & 48.46 & 83.16 & 38.73 & 82.36 &  \\

& G@Ens$_{cov}$  &     - & 86.83 & - & 86.05 & - & 85.87 & - & 84.52 & - & 84.03 & - & 83.59 & - & 82.94 & 17.05  \\

\midrule
\multirow{3}{*}{\textbf{TriviaQA}} & Given (G) & 96.67 & 67.31 & 86.89 & 65.05 & 82.54 & 63.82 & 68.81 & 60.39 & 61.44 & 58.39 & 55.11 & 56.48 & 47.12 & 54.03 & 
   \\

& Ens & 98.01 & 67.58 & 89.88 & 65.71 & 87.99 & 65.15 & 72.31 & 60.95 & 65.0 & 59.13 & 58.47 & 56.9 & 49.67 & 54.38  & 
  \\

& G@Ens$_{cov}$  &     - & 67.64 & - & 65.76 & - & 65.3 & - & 61.38 & - & 59.25 & - & 57.55 & - & 54.94 & 9.5
  \\

\midrule
\multirow{3}{*}{\textbf{NQ}} & Given (G) & 92.37 & 63.78 & 79.04 & 59.99 & 74.87 & 58.77 & 60.60 & 53.51 & 54.03 & 51.00 & 47.94 & 48.31 & 41.70 & 45.27 & 
  \\

& Ens & 94.59 & 64.35 & 83.46 & 60.82 & 81.32 & 60.16 & 64.83 & 54.7 & 58.05 & 52.17 & 51.8 & 49.8 & 44.33 & 46.31 & 
  \\

& G@Ens$_{cov}$  &     - & 64.43 & - & 61.31 & - & 60.79 & - & 55.03 & - & 52.61 & - & 50.01 & - & 46.82 & 13.4  \\

\midrule
\multirow{3}{*}{\textbf{DROP}} & Given (G) & 95.74 & 88.46 & 81.17 & 87.38 & 76.11 & 87.33 & 62.34 & 86.23 & 53.69 & 85.38 & 48.77 & 84.45 & 43.05 & 85.01 &   \\

& Ens & 97.6 & 88.48 & 85.63 & 87.72 & 83.17 & 87.28 & 65.34 & 86.15 & 56.55 & 85.65 & 50.37 & 84.54 & 44.78 & 84.99 &   \\

& G@Ens$_{cov}$  &     - & 88.47 & - & 87.72 & - & 87.52 & - & 86.05 & - & 85.63 & - & 84.54 & - & 84.84 & 1.57
  \\

\midrule
\multirow{3}{*}{\textbf{DuoRC}} & Given (G) &  97.20 & 68.68 & 87.87 & 66.41 & 84.21 & 65.82 & 71.09 & 62.42 & 64.16 & 61.47 & 57.16 & 59.91 & 50.03 & 58.46 &   \\

& Ens & 98.0 & 68.86 & 90.34 & 67.11 & 88.61 & 66.84 & 73.82 & 63.36 & 66.96 & 62.19 & 59.96 & 60.78 & 51.57 & 58.4 & \\

& Original@cov &     - & 68.91 & - & 67.18 & - & 66.69 & - & 63.18 & - & 61.79 & - & 60.07 & - & 58.91 & -1.69 \\

\midrule
\multirow{3}{*}{\textbf{TBQA}} & Given (G) &  94.34 & 67.14 & 80.9 & 63.32 & 75.65 & 61.92 & 57.49 & 56.02 & 49.63 & 52.14 & 41.45 & 51.04 & 34.07 & 50.00 & \\

& Ens & 95.94 & 67.55 & 84.3 & 64.17 & 81.1 & 63.33 & 62.28 & 56.94 & 53.96 & 54.25 & 45.78 & 52.33 & 37.72 & 51.15 &   \\

& G@Ens$_{cov}$  &     - & 67.45 & - & 64.33 & - & 63.38 & - & 57.05 & - & 54.38 & - & 52.03 & - & 50.53 & -6.93 \\

\bottomrule
\end{tabular}
}
\caption{\textbf{Performance of Ensembling (Mean) as a post-abstention method} for selected abstention thresholds. 
The QA model is trained using SQuAD training data and evaluated on SQuAD (in-domain) and 10 out-of-domain datasets.
For each dataset, we provide three rows:
first row (`\textit{Given}') shows the coverage and risk values of the given selective prediction system at different abstention thresholds,
second row (`\textit{Ens}') shows the coverage and risk after applying Ens on abstained instances of the given system, and
third row (`\textit{G@Ens}$_{cov}$') shows risk of the given system at the coverage achieved by \textit{Ens}.
For the post abstention method to be effective, risk in the second row should be less than that in the third row and the magnitude of difference corresponds to the improvement.
The last column corresponds to the overall improvement aggregated over all confidences ranging from 0 to 1 at an interval of 0.02.
$\downarrow$ and $\uparrow$ indicate that lower (risk) and higher (coverage, risk improvement) values are better respectively.}
\label{tab:Ensembling_perf}
\end{table*}

\section{Other Post-Abstention Techniques}
\label{sec_appendix_dicussion}
Asking clarifying questions to the user in order to get information about the question has started to received considerable research attention in conversational, web search, and information retrieval settings \cite{aliannejadi-etal-2021-building, aliannejadi2020convai3, Zamani2020GeneratingCQ, zhang-etal-2020-dialogpt, mimics}. These techniques can be leveraged/adapted for the post-abstention task.

Test-time adaptation is another promising research area in which the model is adapted at test-time depending on the instance. This is being studied in both computer vision \cite{chen2022contrastive} and language processing \cite{wang-etal-2021-efficient-test, banerjee-etal-2021-self}.

Cascading systems in which stronger and stronger models are conditionally used for inference is also an interesting avenue to explore with respect to Post-Abstention \cite{varshney-baral-2022-model, li-etal-2021-cascadebert-accelerating, varshney2022can}.

\section{Coverage 100\% for Human Intervention Methods}
We believe that the ability to identify situations when there is no good answer in the top N returned candidates is a very difficult task (for the humans also) and it requires even more cognitive skills than just selecting the best answer from the provided answer candidates. 
Because of this reason, the coverage is 100\%.

\section{Comparison with Other Selective Prediction Methods}
In this work, we presented a new QA setting and studied the performance of several baseline methods for this task. The focus of this work is on studying the risk improvement that can be achieved in this problem setup. We consciously do not pitch the approaches for this task as competitors of the existing selective prediction approaches. In fact, these approaches are \textbf{complimentary} to the selective prediction approaches. A post-abstention method can be used with any selective prediction method as the first step.

\end{document}